\documentclass[letterpaper, 10 pt, conference]{ieeeconf}  

\IEEEoverridecommandlockouts                              
\usepackage{graphics} 
\usepackage{epsfig} 
\usepackage{times} 
\usepackage{amsmath} 
\usepackage{amssymb}  
\usepackage{color}
\usepackage{verbatim}
\usepackage{comment}
\usepackage{fancyhdr}
\usepackage{xcolor}
\usepackage{txfonts}
\usepackage[algo2e,ruled,vlined]{algorithm2e}
\let\labelindent\relax
\usepackage{enumitem}
\usepackage{optidef}
\usepackage{bbold}
\usepackage{cite}
\usepackage{float}
\usepackage{todonotes}

\usepackage{threeparttable}

\usepackage{hyperref}
\usepackage{algorithm,algpseudocode}
\usepackage{mathrsfs}

\hypersetup{
    colorlinks=true,
    linkcolor=blue,
    filecolor=magenta,      
    urlcolor=cyan,
    pdftitle={Overleaf Example},
    pdfpagemode=FullScreen,
    }

\usepackage[font=small]{caption}

\usepackage[compact]{titlesec}

\title{Cross-Embodiment Robotic Manipulation Synthesis via Guided Demonstrations through CycleVAE and Human Behavior Transformer}

\author{Apan Dastider$^{1}$, Hao Fang$^{1}$ and Mingjie Lin$^{1}$ 
\thanks{$^{1}$Department of Electrical and Computer Engineering, University of Central Florida, Orlando, FL, 32816, USA (E-mail: apan.dastider@ucf.edu))}}

\begin{document}

\setlength{\textfloatsep}{10pt}

\maketitle
\thispagestyle{empty}
\cfoot{\thepage}
\renewcommand{\headrulewidth}{0pt}
\pagestyle{empty}
\cfoot{\thepage}
\setlength{\textfloatsep}{3pt}
\setlength{\floatsep}{3pt}

\begin{abstract}
Cross-embodiment robotic manipulation synthesis for complicated tasks is challenging, partially due to the scarcity of paired cross-embodiment datasets and the impediment of designing intricate controllers. Inspired by robotic learning via guided human expert demonstration, we here propose a novel cross-embodiment robotic manipulation algorithm via CycleVAE and human behavior transformer. First, we utilize unsupervised CycleVAE together with a bidirectional subspace alignment algorithm to align latent motion sequences between cross-embodiments. Second, we propose a casual human behavior transformer design to learn the intrinsic motion dynamics of human expert demonstrations. During the test case, we leverage the proposed transformer for the human expert demonstration generation, which will be aligned using CycleVAE for the final human-robotic manipulation synthesis. We validated our proposed algorithm through extensive experiments using a dexterous robotic manipulator with the robotic hand. Our results successfully generate smooth trajectories across intricate tasks, outperforming prior learning-based robotic motion planning algorithms. These results have implications for performing unsupervised cross-embodiment alignment and future autonomous robotics design. Complete video demonstrations of our experiments can be found in \url{https://sites.google.com/view/humanrobots/home.}
            
\end{abstract}
\section{INTRODUCTION}

Knowledge transfer and domain adaptation between systems that differ in geometrical configurations and parametric representations impose rudimentary learning challenges. Although knowledge distillation between different systems is a promising direction for knowledge transfer, it suffers from dissimilar morphologies and dynamical kinematics, which stop achieving the accurate levels of abstraction and universal representations in both state-space and actions-space\cite{transferLearn, yang2024pushing}. Meanwhile, learning robotic manipulation skills and smooth motion synthesis for many complicated tasks is challenging because of the scarcity of diverse, guided datasets\cite{firoozi2023foundation} and the impediments of designing sophisticated robotic controllers\cite{zhang2024catch,dastider2024retro}. Additionally, the accumulation of control sequence samples may become difficult for complex robotic systems due to the requirements of fine-crafted robotic controllers. To this end, more advanced algorithms should be developed urgently to address the following requirements: 1) the collection of cross-embodiment robotic manipulation dataset is efficient; 2) identifying corresponding behaviors and control policies between the cross-embodiment robotic systems requires learning the generalizable mapping functions\cite{wang2024cross}; 3) the algorithm ensures robotic manipulation synthesis end-to-end with smooth and applicable trajectory guarantees.

\begin{figure}                                     
    \centering                                             
    \includegraphics[width=0.95\linewidth]{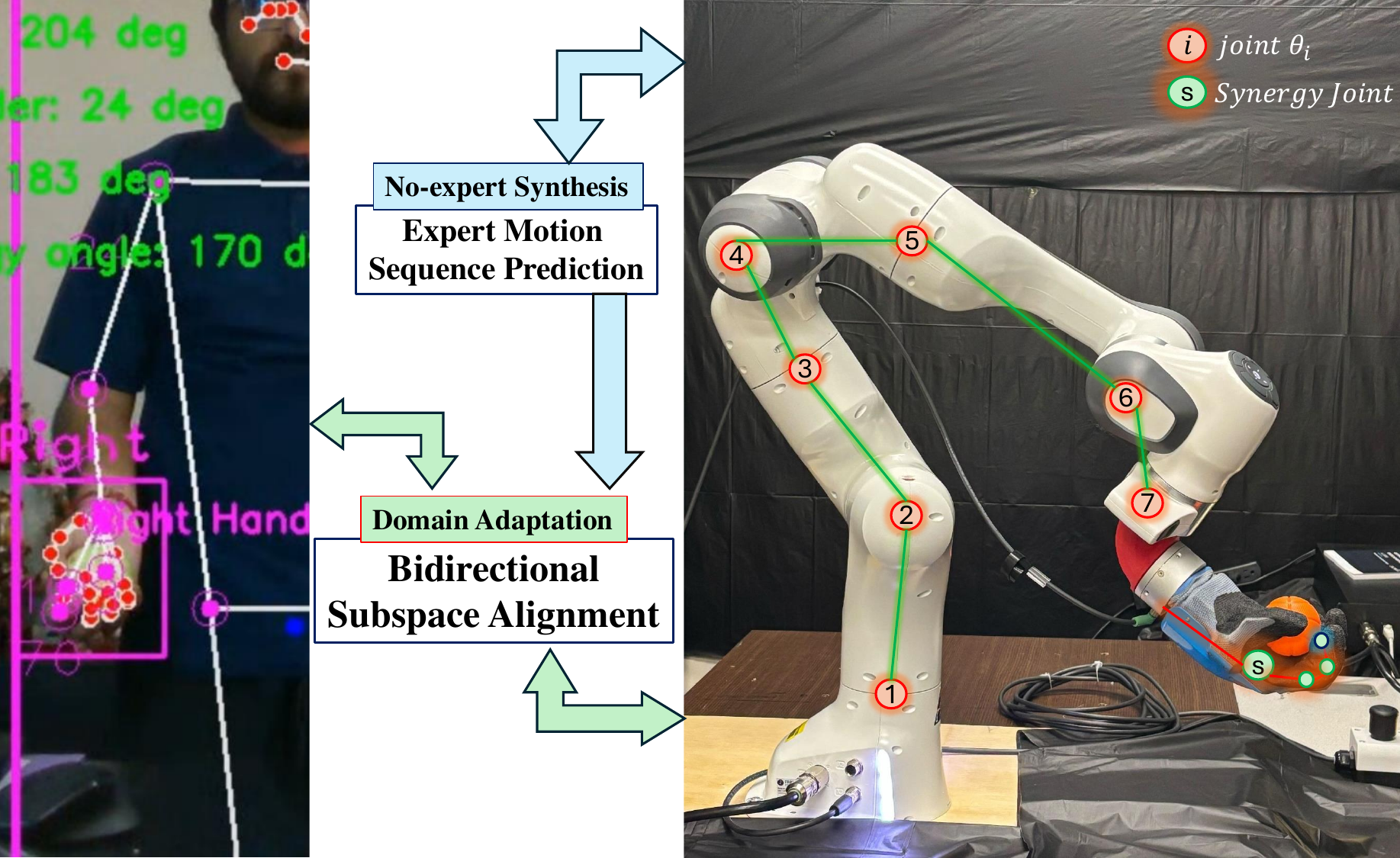}  
    \caption{Human expert demonstration (left) and Robotic motion (right). The knowledge distillation between two motions through our developed bidirectional subspace alignment method (middle).} 
    \label{fig:title}                                   
\end{figure}

Our primary concentration on this work focuses on knowledge transfer and domain adaptation to synthesize safe and smooth robotic motion planning trajectories guided by expert demonstrations for solving intricate tasks such as ball-tossing and catching with
a dexterous robotic manipulation using the robotic anthropomorphic hand. To achieve the above goal and address three prior requirements, we develop a novel cross-embodiment robotic manipulation algorithm for the trajectory synthesis via guided demonstrations using CycleVAE and human behavior transformer.  
First, we use human expert demonstration for data collection (figure~\ref{fig:title} left). Thus, we can accumulate the joint motion, velocity profile, and torque calculation of human arm joints from video processing in real time. On the contrary, collecting robotic dataset is time-consuming because it requires the well-design of hand-crafted robotic controller to ensure proper synchronization and coordination between robotic arm and hand. With the imbalanced and unpaired datasets collected from human and robotic systems, we propose CycleVAE\cite{jha2018disentangling} with a bidirectional subspace alignment objective in an embodiment-agnostic fashion. The CycleVAE aligns the knowledge from two different domains into a common latent representation, which later can be used to guide the learner domain for the generation of robotic manipulation trajectories (figure~\ref{fig:title} middle). We further innovatively propose a causal human behavior transformer to exclusively learn the internal human behavior dynamics, which can function as an artificial human for even swiftly synthesizing thousands of human expert demonstrations in seconds. We tested our proposed algorithm on 7 DoFs Franka Emika Panda Robotic Manipulator attached with QB Soft hand 2 (figure~\ref{fig:title} right). Our experimental results successfully and smoothly generate trajectories across intricate tasks, surpassing conventional and deep learning-based robotic motion planning algorithms. These results carry significant developments and influential directions for performing unsupervised knowledge distillation among varying domains and alleviating robotic motion generation autonomously for many intricate tasks.
 
\begin{figure*}[t]                                     
    \centering                                             
    \includegraphics[width=0.95\linewidth]{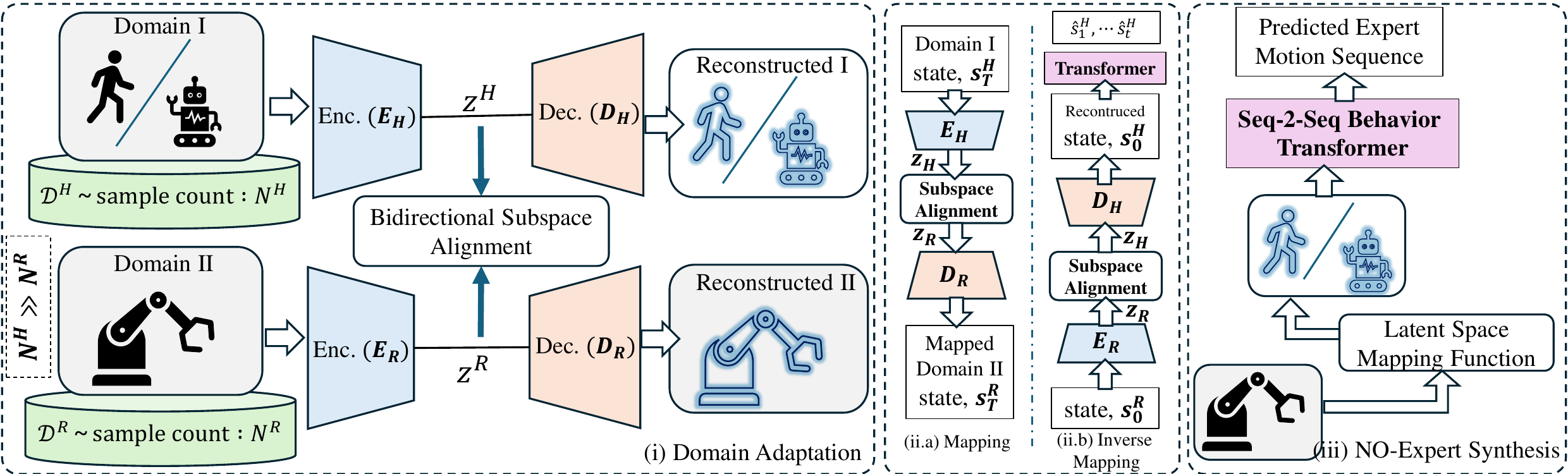} 
    
    \caption{Overall diagram of our contributions: (i) domain adaptation via bidirectional subspace alignment algorithm; (ii.a) Mapping flow from human demonstration to robotic motion synthesis, (ii.b) When unavailable demonstrations, bidirectional mapping from robot initial state to expert state and expert future motion prediction;   (iii) No-expert trajectory synthesis using causal human behavior transformer}
    \label{fig: overall}                                   
\end{figure*}


\section{RELATED WORK}
\subsection{Cross-embodied knowledge transfer}
Cross-embodied domain adaptation, such as teacher robot to student robot and human-to-robot knowledge transfer, has appeared as an emerging field that focuses on reducing the gap between human motion analysis and robotic learning through observing expert demonstrations. For instance, \cite{jain2024vid2robot} proposes the cross-attention mechanism with contrastive losses to enable robots to complete assigned manipulation tasks by observing human demonstrations and learning a unified representation by using paired human video and robot trajectory datasets. \cite{mendonca2023structured} builds a structured shared human-robot action space leveraging visual affordances from video demonstrations and later finetunes the world model by collecting robot samples from the affordance space in an unsupervised way. Another work \cite{wang2024cross} learns the mapping of control policies between robotic manipulators with different morphologies by building a shared latent representation of the original state space and action space of robotic manipulators through utilizing a generative adversarial training mechanism. \cite{song2021attention} proposes the Human-to-Robot Attention Transformer (H2R-AT), which identifies and corrects incorrect robotic manipulations at an early stage by transferring human verbal instructions. \cite{Li2022MetaImitationLB} proposes A-CycleGAN, an advanced generative model that translates human demonstration videos into real-world robot demonstrations by training the meta-policy in a compact latent space. \cite{robotube} develops useful and reproducible benchmark RT-sim as the simulated twin environments to facilitate household robotic manipulation tasks from human videos and ease the process of data collection. 
\cite{eze2024learning} is a comprehensive review paper that discusses various cutting-edge robotic manipulation learning algorithms by observing human demonstrations. Although there is a surge of development in cross-embodied policy transfer learning, but these methods either require a paired dictionary of demonstrations or the training pipeline is very resource expensive due to image-based feature extractions and learning. 

\subsection{Transformer-based Robotic manipulation algorithm}

\cite{brohan2022rt} proposes Robotics Transformer (RT1), a novel architecture to learn robotic control utilizing both image features and natural language task description by training a transformer model along with pre-trained ImageNet and token learner module to produce descretized action tokens. \cite{chen2021decision} reformulates deep reinforcement learning problems as a sequence learning problem using a transformer model that predicts future actions based on past state information, computed actions, and gained rewards. \cite{wang2024hpt} introduces Heterogeneous Pre-trained Transformers (HPT), a shareable large trunk of policy network to learn embodiment-agnostic shared representation by aligning proprioception and vision inputs into a sequence of tokens and processing those tokens to output control values. Another work \cite{dasari2024ditpi} integrates the advantages of the transformer model and diffusion model into building a unique architectural design for transformer diffusion policy learning, which can generate actions for diverse tasks on different embodiments without requiring extensive parameter tuning.  

\begin{figure*}                                     
    \centering                                             
    \includegraphics[width=0.9\linewidth]{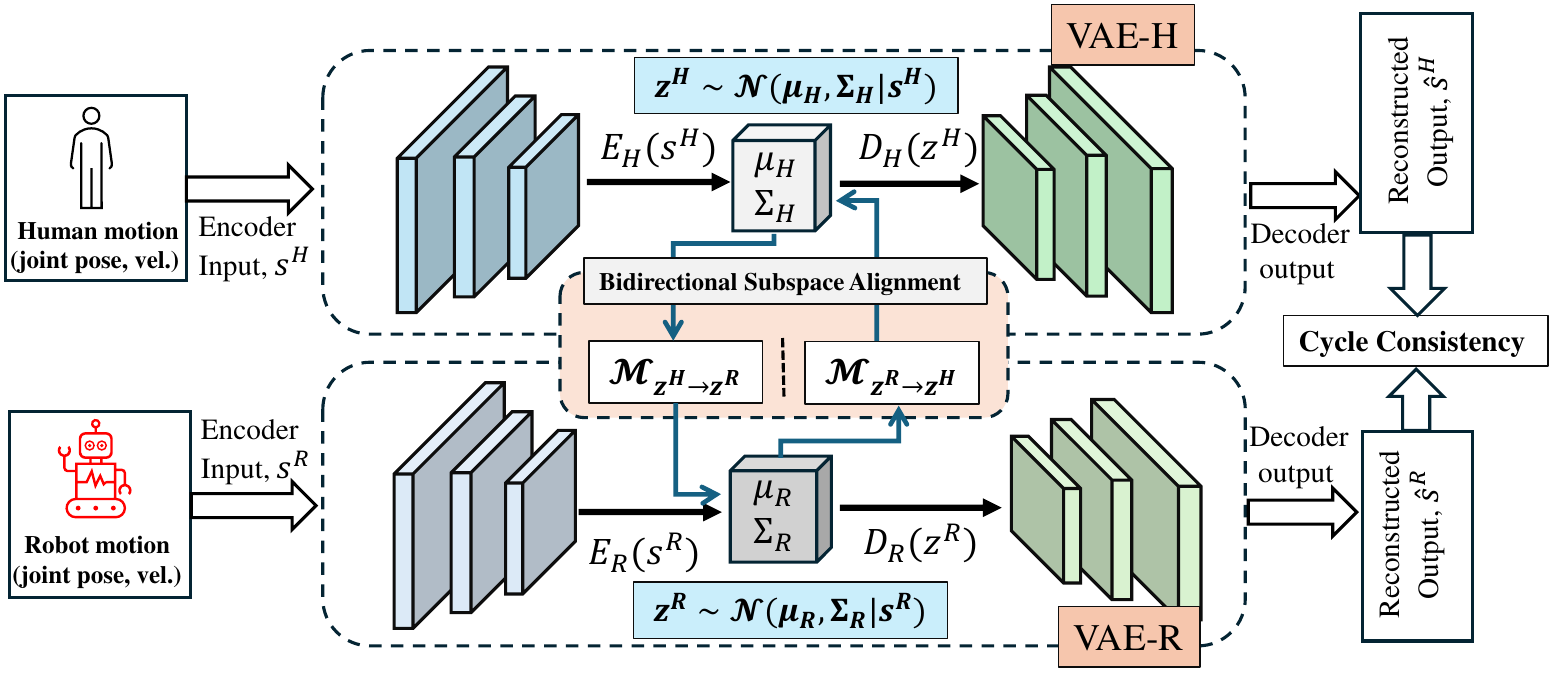}  
    \caption{Our proposed cycle-VAE architecture for bidirectional alignment of human motion and robot motion. We build two VAEs to learn the latent representation of two motion modalities. Further, we apply cycle consistency loss to ensure the alignment is bidirectional. }%
    \label{fig: cycleVAE}                                   
\end{figure*}

\section{METHODS}
\subsection{Problem formulation}

We consider the general domain shift problems where two robotic systems differ in parametric representation and dynamics model. The dynamics of a robotic manipulator are governed by the Euler-Lagrange equations in joint space that describe how joint torques influence motion and generate smooth trajectories with given joint positions, velocities, and accelerations. The equation for $n$ degrees-of-freedom (DoFs) robotic manipulator is represented as:
\begin{equation}
    M(q)\ddot{q}+C(q,\dot{q})\dot{q}+G(q) = \tau,   
\end{equation}
where $q\in\mathbb{R}^n$ is the vector of joint positions, $\dot{q}\in\mathbb{R}^n$ is the vector of joint velocities, $\ddot{q}\in\mathbb{R}^n$ is the vector of joint accelerations, $M(q)\in\mathbb{R}^{n\times n}$ is the mass matrix, $C(q,\dot{q})\in\mathbb{R}^{n}$ is the Coriolis and centrifugal forces, $G(q)\in\mathbb{R}^n$ is the gravitational torque and $\tau\in\mathbb{R}^n$ is the actuated joint torque. Now, let us assume that there exists two different robotic manipulators $R_H$ and $R_R$ differing in degrees of freedom and geometrical configurations. $R_H$ contains $m$ DoFs and $R_R$ has $n$ DoFs for manipulation in joint space, and $n\neq m$. The equation of motion will be given by 

for manipulation system $R_R$, we have
\begin{equation}
M_R(q_R)\ddot{q}_R+C_R(q_R,\dot{q}_R)\dot{q}_R+G_R(q_R) = \tau_R. 
\label{Robot eq}
\end{equation}

for manipulation system $R_H$, we have
\begin{equation}
M_H(q_H)\ddot{q}_H+C_H(q_H,\dot{q}_H)\dot{q}_H+G_H(q_H) = \tau_H. 
\label{Human eq}
\end{equation}
The one-to-one action mapping and direct control strategy transfer seems unrealistic and error-prone since the parameter space and matrices in Eq.\eqref{Robot eq} and Eq.\eqref{Human eq} are totally dissimilar in dimension and representation. The mismatches among $M(\cdot), C(\cdot,\cdot), G(\cdot)$ can be mathematically solved by, 
\begin{align}
\begin{split}
    \Delta M_{HR} &= M_H(\cdot)-G_{M_R}(M_R(\cdot)), \\
    \Delta C_{HR} &= C_H(\cdot,\cdot)-G_{C_R}(C_R(\cdot,\cdot)), \\
    \Delta G_{HR} &= G_H(\cdot)-G_{G_R}(G_R(\cdot)).
\end{split}
\end{align}
where $\Delta M_{HR},\;\Delta C_{HR},\;\Delta 
G_{HR}$ are respectively the differences between mass matrices, Coriolis and centrifugal matrices, and gravitational torque matrices. Additionally, $G_{M_R}, G_{C_R}, G_{G_R}$ are the calculated transformation matrices to transform respectively $M_R, C_R, G_R$ to the same configuration space of the another system $R_H$. But, accurate calculation of these transformation matrices imposes serious challenges and requires huge mathematical derivations. Moreover, $M_H,(\cdot),C_H(\cdot,\cdot),G_H(\cdot)$ may not be available if knowledge and learning are shared between a human expert system, $R_H$ and a robotic learner, $R_R$. 

The control mapping and torque control calculation for $R_R$ guided by domain $R_H$ can be solved by, 
\begin{equation}
    \tau_R = \tau_H + \Delta \tau_{HR}(q,\dot{q},\ddot{q}),
\end{equation}
where, $\Delta \tau_{HR}(q,\dot{q},\ddot{q})=\Delta M_{HR}(q)\ddot{q}+\Delta C_{HR}(q,\dot{q})\dot{q}+\Delta 
G_{HR}(q)$. The mapping complexity is then proportional to the effort required to compute and adapt the torque correction term: 
\begin{equation}
    D_\tau = \int\int\int||\Delta\tau_{HR}(q,\dot{q},\ddot{q})||^2 dq .d\dot{q}. d\ddot{q}.
\end{equation}
Thus, computations of the above mathematical formulations are analytically complicated and the matrix $M_H, C_H, G_H$ are not always available for systems such as human arm movement. In contrast, the data-driven approach seems promising for learning complex mapping functions among heterogeneous embodied domains \cite{wang2024cross, jain2024vid2robot}. However, the well-established domain adaptation methods often require paired data sample dictionaries to facilitate knowledge sharing, which is not practical for many intricate robotic tasks. Different robotic systems have their own space and time complexities. Here, we aim to learn mapping functions in an unsupervised manner where both datasets are collected from two systems. Those two systems do not contain paired trajectory samples (i.e., they are dissimilar in geometrical configurations and joint-space representation) but are under the assumption that they both can accomplish similar tasks. Below, we propose an innovative cycle-VAE architecture to achieve this goal (also see figure~\ref{fig: cycleVAE}). 

\subsection{Cycle-VAE for the bidirectional alignment of Human and Robot}

\begin{figure}                                     
    \centering                                             
    \includegraphics[width=\linewidth]{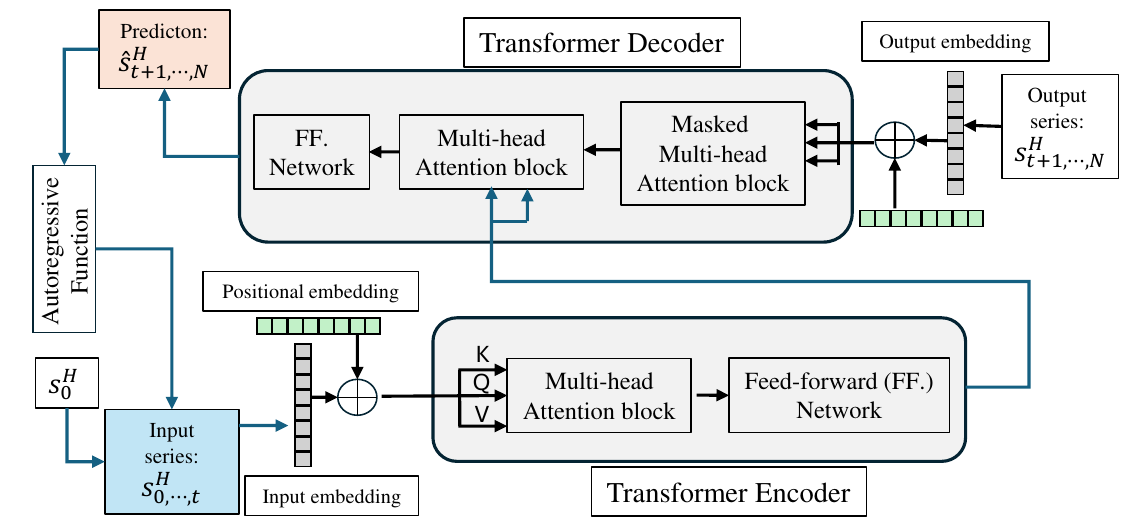}  
    \caption{
    The architecture of our proposed Human Behavior Transformer. We use the masked multi-head attention block in the transformer decoder to ensure causal modeling. Our human behavior transformer is trained in auto-regressive way to predict future human behaviors.}%
    \label{fig:human_transformer}                                   
\end{figure}

Consider that we had human demonstration data ($s^H_0\in \mathbb{R}^{d^H}$), and we want to map it into a feasible robotic joint trajectory data ($s^R_0\in \mathbb{R}^{d^R}$) through a mapping function $\mathcal{M}_{H\rightarrow R}$ ($s^R_0:=\mathcal{M}_{H\rightarrow R}(s^H_0)$). Meanwhile, we also want to have a reverse representation mapping functional $\mathcal{M}_{R\rightarrow H}$ which will output $s^H_0:=\mathcal{M}_{R\rightarrow H}(s^R_0)$. To alleviate these challenges, we develop a Cycle Variational Autoencoder (CycleVAE) with bidirectional latent space mapping between human and robotic data (see figure~\ref{fig: cycleVAE}). The bidirectional alignment in latent space appears with following advantages. Firstly, bidirectional mapping instead of unidirectional ensures the consistent alignment between two domains in latent space representation. Secondly, our bidirectional learning objective allows both domains to stimulate motion dynamics sharing to learn better knowledge distillation. Third, the proposed bidirectional map can also contribute to synthesize robotic motion trajectories when expert demonstrations of guidance are absent, and we aim to achieve self-reliant motion generation for dynamic tasks (see next section, also figure~\ref{fig:human_transformer}).

Given human demonstration data $s^H\in \mathbb{R}^{d^H}$ and robotic trajectory data $s^R\in \mathbb{R}^{d^R}$, where $d^H \neq d^R$ and $d^H>d^R$, our Cycle-VAE consists of coupled VAEs via a latent bidirectional mapping. Specifically, 
\begin{itemize}
    \item The first VAE for encoding human demonstration data (denoted as VAE-H): The VAE-H architecture focuses on learning a latent representation, $z^H$ of the high-dimensional human motion data through VAE encoder ($E_{H}$). The learned latent representation follows Gaussian distribution $\mathcal{N}(\mu^H,\Sigma^H)$, i.e., $z^H \sim f_H(z^H|s^H):= \mathcal{N}(\mu^H,\Sigma^H)$. The decoder part, $D_{H}$ serves the purpose of reconstructing the human demonstration data $\hat{s}^H$, which will be used for later bidirectional mapping. In general, we have 
    \begin{equation*}
        z^H = E_{H}(s^H), \;\; \hat{s}^H = D_{H}(z^H).
    \end{equation*}
    \item The second VAE is designed for encoding robotic trajectory data (denoted as VAE-R). Similar to the first VAE-H, the VAE-R model encodes the high-dimensional robot data into a latent representation $z^R$ using encoder $E_{R}$, and the decoder $D_{R}$ focuses on the reconstruction $\hat{s}^R$. In short, we have 
    \begin{equation*}
        z^R = E_{R}(s^R), \;\; \hat{s}^R = D_{R}(z^R).
    \end{equation*}
    \item Latent subspace bidirectional mappings $\mathcal{M}_{z^H\rightarrow z^R}$ and $\mathcal{M}_{z^R\rightarrow z^H}$: This is the key in our cycle-VAE, which guarantees consistent bidirectional mapping between two modalities $z^R$ and $z^H$ at their latent representations. We include cycle-consistency loss functions to ensure the latent representative distribution alignment. After training, the mapping $\mathcal{M}_{z^H\rightarrow z^R}$ and $\mathcal{M}_{z^R\rightarrow z^H}$ can work efficiently for knowledge transfer and domain adaptation through the low-dimensional latent space.   
\end{itemize}
To train our Cycle-VAE, we use the following objective loss function, which consists of multiple well-crafted and modified loss terms to ensure consistent knowledge transfer, multi-modal domain adaptation, and accurate bidirectional mapping. Specifically, 
\begin{enumerate}[label=(\alph*)]
    \item both VAE-H and VAE-R should realize good reconstruction and maintain Gaussian regularization.
    \begin{align}
\begin{split}
    L_\text{VAE-H} &= \mathbb{E}[||s^H-\hat{s}^H||^2]+D_{KL}(f_H(z^H|s^H)||p(z^H)), \\
    L_\text{VAE-R} &= \mathbb{E}[||s^R-\hat{s}^R||^2]+D_{KL}(f_R(z^R|s^R)||p(z^R)), 
\end{split}
\end{align}
where, $D_{KL}(\cdot||\cdot)$ represents the $KL$ divergence. 
    \item we next introduce cycle loss~(inspired from \cite{zhu2017unpaired}) to enforce bi-directional consistency:
    \begin{equation}
    \begin{split}
        L_{cycle} = \mathbb{E}[||D_H(E_R(D_{R}(E_{H}(s^H))))-s^H||^2]+\\ \mathbb{E}[||D_R(E_H(D_{H}(E_{R}(s^R))))-s^R||^2].
        \end{split}
    \end{equation}
    This loss function enforces the consistent learning such that human data can be recovered while transforming through $\mathcal{M}_{H\rightarrow R}(\cdot)$ and $\mathcal{M}_{R\rightarrow H}(\cdot)$ functions. Similar enforcement is for the robotic data. With this loss function, the model learns to focus on the important features that can represent the core relevant motion patterns and ignore irrelevant noise transformations.
    \item Although the above two loss functions handle the challenge of knowledge transfer between heterogeneous domains, we added another loss to address the potential challenge, where the one-to-one motion translation dictionary is not always available within our dataset. Typically, it is time-consuming to collect large corresponding datasets. So, we enforce additional alignment for the learning latent representation $\mathcal{N}(\mu^H,\Sigma^H)$ and $\mathcal{N}(\mu^R,\Sigma^R)$ using Maximum Mean Discrepancy (MMD) and Covariance Eigenvector Matching. The MMD aligns the marginal latent space distributions of human motion and robot motion by following the principles of first-order statistics: 
    \begin{equation}
        L_\text{mean} = ||\mu^H-\mu^R||_2^2.
    \end{equation}
    Next, we perform Eigenvector-based matching for covariance alignment by the minimization of the Frobenius norm of the difference between top-$k$ eigenvectors\cite{luo2020unsupervised}:
    \begin{equation*}
        \Sigma^H = Q^H\Lambda^H(Q^H)^T, \;\; \Sigma^R = Q^R\Lambda^R(Q^R)^T, \\
    \end{equation*}
    \begin{equation*}
        Q^H_{(k)} = [q^H_1,q^H_2,\cdots,q^H_k], \;\; Q^R_{(k)} = [q^R_1,q^R_2,\cdots,q^R_k],
    \end{equation*}
    \begin{equation}
        L_\text{covariance} = ||Q^H_{(k)}-Q^R_{(k)}||^2_F,
    \end{equation}
    where $Q$ is the Eigenvectors of $\Sigma$. This loss function preserves geometrical relationships and ensures feature eigen-directional matching in latent human and robotic space. 
\end{enumerate}
To this end, the final loss function can be constructed as:
\begin{equation}
    L = L_\text{VAE-H}+L_\text{VAE-R}+\lambda_1 L_{cycle}+\lambda_2 (L_\text{mean}+L_\text{covariance}),
\end{equation}
where $\lambda_1, \lambda_2$ are hyperparameters controlling the trade-off between reconstruction accuracy, cycle consistency, and latent space alignment.
\subsection{Human behavior transformer}
Our proposed cycle-VAE can serve as the bidirectional alignment foundation for human manipulation and robotic trajectories. However, compared to the dense and feasible space of human expert manipulation, we usually can only collect finite human expert demonstrations in practice, which is relatively sparse. To address the above data sparsity issue, we propose a \textit{Human Behavior Transformer} as a generative model for human expert manipulation. Specifically, our human behavior transformer is a causal sequence-to-sequence transformer model, which can predict future human motions based on past observations. Consider a sequence of human motion data $\mathcal{S}^H_{0, \cdots, t}={ \{s^H_0, s^H_1, \cdots, s^H_t \}}$, where each time stamp $s^H_t\in\mathbb{R}^{d^H}$ represents the human joint measures (e.g., velocity), we can train a causal transformer model for the future step prediction $\mathcal{\hat{S}}^H_{t+1, \cdots, N}={\{ s^H_{t+1}, s^H_{t+2}, \cdots, s^H_N \}}$, where $N$ indicates the total length of the manipulation trajectory. 
\begin{equation}
   \mathcal{S}^H_{t+1, \cdots, N}=f_\Gamma (\mathcal{S}^H_{0, \cdots, t}).
\end{equation}
Here $f_\Gamma$ is a causal transformer model parameterized by $\Gamma$, which can be trained in an auto-regressive fashion. The detailed structure of our \textit{Human Behavior Transformer} can be found in figure~\ref{fig:human_transformer}. 
Introducing the \textit{Human Behavior Transformer} is the second innovative contribution in our work. First, the well-trained generative transformer can be treated as an artificial human expert, which can generate various feasible human expert demonstrations. Second, regarding the data collection (see next section of the experimental setup), our \textit{Human Behavior Transformer} essentially generates data faster than the actual execution of human experts, which provides great computational advantages (see results section for details).

To this end, we complete our algorithm design. An overall framework of our algorithm can be found in algorithm~\ref{Algorithm 1}.
\begin{algorithm}
	\caption{CycleVAE Model}
	\label{Algorithm 1}
        \textbf{--------TRAINING--------}\\
        \small
	\textbf{Input:} Human Data, $\mathcal{D}^H:=\{\tau_i\}_{i=1}^{N^H}$, Robot data, $\mathcal{D}^R:=\{\tau_j\}_{j=1}^{N^R}$ ,
    \;\;\;Bidirectional Model, $f_\Theta(\cdot)$, Training Steps, $T$\\
        \textbf{Output:} Trained Model, $\hat{f}_\Theta(\cdot)\;\text{where} \;\Theta:=\{\theta_1,\theta_2,\theta_3,\theta_4,\theta_5,\theta_6\}$ \\
        \textbf{Initialize:} $E_H^{\theta_{1}}, E_R^{\theta_{2}}, D_H^{\theta_{3}}, D_R^{\theta_{4}}, 
        \mathcal{M}_{z^H\rightarrow z^R}^{\theta_{5}},
        \mathcal{M}_{z^R\rightarrow z^H}^{\theta_{6}}$\\
  \SetKwProg{Fn}{Function}{}{}
\Fn {\text{train($\mathcal{D}^H,\mathcal{D}^R, f_\Theta(\cdot), T$):}}{
		\While{not converged}{
        \scriptsize \texttt{\# sample batch of human and robot traj.}\\
        \small
        $\tau^H_n\sim\mathcal{D}^H,\;\tau^R_n\sim\mathcal{D}^R$; \#\;$\tau^H_i:=\{s^H_k\}_{k=0}^t,\;\tau^R_j:=\{s^R_k\}_{k=0}^t$\\
        \scriptsize \texttt{\# Encode into Latent Space}\\
        \small
        $z^H = E_H^{\theta_1}(s^H),\;z^R = E_R^{\theta_2}(s^R)$; \\$\#\;z^H\sim \mathcal{N}(\mu^H,\Sigma^H),\;\;z^R\sim \mathcal{N}(\mu^R,\Sigma^R)$\\
        \scriptsize \texttt{\# Reconstructed into Original Space}\\
        \small
        $\hat{s}^H = D_H^{\theta_3}(z^H),\;\hat{s}^R = D_R^{\theta_4}(z^R)$;\\
        \scriptsize \texttt{\# Compute all loss functions}\\
        \small
        $L_{VAE-H} = ||s^H-\hat{s}^H||^2+D_{KL}(\cdot|\cdot)$\\
        $L_{VAE-R} = ||s^R-\hat{s}^R||^2+D_{KL}(\cdot|\cdot)$\\
        $L_{cycle} = ||D_H(E_R(D_{R}(E_{H}(s^H))))-s^H||^2+\\||D_R(E_H(D_{H}(E_{R}(s^R))))-s^R||^2$\\
        $L_{align} = L_{MMD}+L_{covariance}$
        \\
        \scriptsize \texttt{\# Total loss and Grad. Update}\\
        \small
        $L_{total}=L_{VAE-H}+L_{VAE-R}+\lambda_1L_{cycle}+\lambda_2 L_{align}$\\
        $\Theta = \Theta+\nabla_\Theta L_{total}$
        }
}
\normalsize
\textbf{--------INFERENCE--------}\\
\small
\textbf{Input:} Expert Demonstration $\tau_D^H$, Trained Model $\hat{f}_\Theta$; \\
\textbf{Output:} Robot Trajectory $\tau^R_G$\\
  \SetKwProg{Fn}{Function}{}{}
\Fn {\text{execute($\tau^H_D, \hat{f}_\Theta$):}}{
    \scriptsize \texttt{\# Mapping into shared latent space}\\
        \small
        $\hat{z}^H=\hat{E}_H^{\theta_1}(s^H_D), \hat{z}^R=\mathcal{M}_{z^H\rightarrow z^R}(\hat{z}^H)$ \\
        \scriptsize \texttt{\# Reconstruct robot trajectory}\\
        \small
        $\hat{\tau}_G^R:=\hat{s}^R_G=\hat{D}_R^{\theta_4}(\hat{z}^R)$\\
        \texttt{EXECUTE($\hat{\tau}_G^R$)}
    }
\end{algorithm}

\begin{figure}                                     
    \centering                                             
    \includegraphics[width=0.7\linewidth]{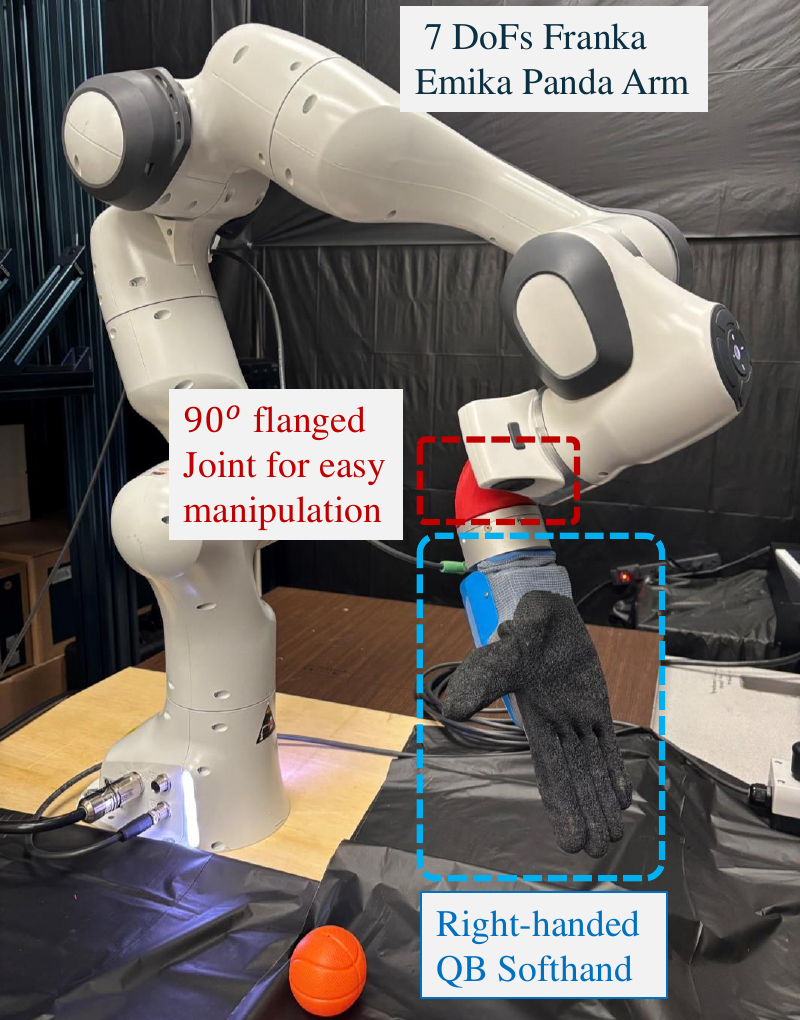}  
    \caption{Ours Hardware Research Robotic Platform: A 7 DoFs Franka Emika Panda Robotic Manipulator attached with QB Softhand 2 Research at $90^o$ joint. A softball in orange color to manipulate}%
    \label{fig:hardware}                                   
\end{figure}

\subsection{Platforms and Experimental setup}
To evaluate our proposed algorithm, we started by building the following evaluation hardware and software platforms: 1) we utilized a high-fidelity research-grade 7-DoF Franka Emika Panda robotic manipulator and assembled it on a fixed tabletop; 2) we used right-handed anthropomorphic five-fingered hands (QB SoftHand2 Research) for replicating human hand motion in ball tossing and catching tasks; 3) we used an Intel RealSense Depth Camera D435i to locate objects and track human motions with in-depth information; 4) we utilized Google Mediapipe tool to real-time record human arm joint motions at each frame. Further, we applied You Only Look Once (YOLO-LITE)~\cite{huang2018yololite} algorithm  for object detection and localization; 5) we applied an extended Kalman Filter (EKF) to estimate the trajectory of moving objects; 6) we transported the $[x,y,z]_b$ information from the camera reference frame to the robot coordinate system using the standard and well-established hand-to-eye calibration~\cite{handtoeye}. To this end, the adaptive trajectory planner utilized predicted trajectory information of the moving object to synthesize the real-time smooth trajectory for grabbing the object. 


With the above platform setup, we replicated the exact model of the Franka Emika Panda Arms with QB SoftHand2 in the Robot Operating System (ROS). The libfranka and Franka ROS established relatively low system latency and high signal-to-noise ratio (SNR) communication protocols for robotic data processing and parallel execution between the simulation and hardware platforms. Further, the QB Softhand2 research hands were operated through serial communication in parallel for object grasping and catching with a synchronized motion.

Our motion task is to perform ball-juggling such that the arm can toss a ball in upwards motion with limited velocity and catch the ball within the limited flying time. 
It is assumed that collecting trajectory samples from the expert domain is easier compared to the robotic operation domain. Therefore, We collected $10,000$ trajectory samples from expert demonstrations and $1,000$ trajectory samples from robotic manipulation. Each trajectory sample consists of $T = 128$ discrete trajectory time steps, representing the temporal states from the
initial state to the final state. Specifically, we sampled high-dimensional robotic arm manipulation trajectories $\{\tau_1,\tau_2, \cdots, \tau_n \}$ from ROS, where each trajectory sample $\tau_i^j$ ($j$ is from ${1,2,3,\cdots 12}$) contains $12$ different float values: $7$ joint values from robotic arm $[\theta^0,\cdots,\theta^6]_R$; $2$ synergy joint values from QB soft robotic hands $[\theta_Q^1,\theta_Q^2]$; 3D information $[x,y,z]_b$ of the thrown object. It is noted that we only exploited the synergy joints-1st synergy of our robotic hands which have illustrated the efficient control of the $19$ self-healing finger joints in prior studies~\cite{dastider2024apex}. On the contrary, we collected human expert demonstrations. Each demonstration sample contains $31$ numeric values - human arm-hand joint values, $\{[x_i, y_i, z_i]_{i=(MCP, PIP; S, E, W)},[\theta_{MCP},\theta_{PIP};\theta_{S}, \theta_{E}, \theta_{W}]\}$ and 3D object information $[x,y,z]_b$ for the ball thrown by a human expert. Specifically, we stored both 3D information and joint values of shoulder $(S)$, elbow $(E)$ and wrist $(W)$ joints from each video frame. We tracked the metacarpophalangeal joint ($MCP-MCP_{ind}, MCP_{mid}$) and proximal interphalangeal joint ($PIP-PIP_{ind}, PIP_{mid}$) joints information of index and middle fingers for only right hand. The above two joint values from two fingers are enough to perform the guided learning and trajectory generation during the ball tossing and catching task. The rest of fingers perfectly synchronizes with the same motion and trajectory profile.

\subsection{Comparison and Evaluation metrics}
We design both comprehensive Monte Carlo computer
simulations and hardware experiments to evaluate our proposed algorithm to validate the knowledge transfer from an expert demonstration and synthesize the motion trajectory for dynamic task completion. We are confident that by illustrating the performance
on the ball toss and catch task with limited velocity and unpredictable flying duration, we
essentially evaluate our algorithms across a wide range of knowledge distillation among heterogeneous domains. This underlying advantage resolves data accumulation difficulties, data scarcity problems and cross-embodied domain adaptation problem. Our algorithm was implemented using the Pytorch library and ran on a Lambda QUAD GPU workstation equipped with an Intel Core-i9-9820X processor and $4$ Nvidia $3080-ti$ GPU machines. For a fair comparison, we also implement other state-of-the-art trajectory generation algorithms where we include Dynamic RRT~\cite{dynamicrrt}, Decision Transformer~\cite{chen2021decision}-sequence modeling with reinforcement learning, DAMON-our prior latent space based path planning algorithm~\cite{Dastider_Damon_IROS_2023}, a transformer-based imitation learning algorithm~\cite{KIM_IL_IROS_2021}, Vid2Robot-a video-conditioned robotic policy learning method~\cite{jain2024vid2robot}. We train each baseline method with same dataset samples we collect for our CycleVAE architecture.
 
We define the following evaluation metrics to quantify the performance effectively. First, we define the succes ratio as  $R_\text{success} = \frac{S}{N} \times 100\%$,
where $S$ counts the successful tasks and $N$ denotes the total number of tasks that we performed. Therefore, larger $R_\text{success}$ means the generated trajectories executed successfully and realized better task completion rates. 
Then, we quantify the smoothness of the executing trajectory of human guided robots by summing up the joint space configuration change in the sampled trajectory as $S_r = \sum_{i=0}^{N-1}||\theta_i - \theta_{i+1}||_2$, where $\theta_i$ is joint space configuration at $\tau_i$. The lesser value means that we achieve a smooth trajectory for motion execution. Finally, we include total computation time, $T_{gen}$, for complete motion generation from our proposed method and baseline methods.

\section{Results}
\begin{figure*}                                   
    \centering                                             
    \includegraphics[width=\linewidth]{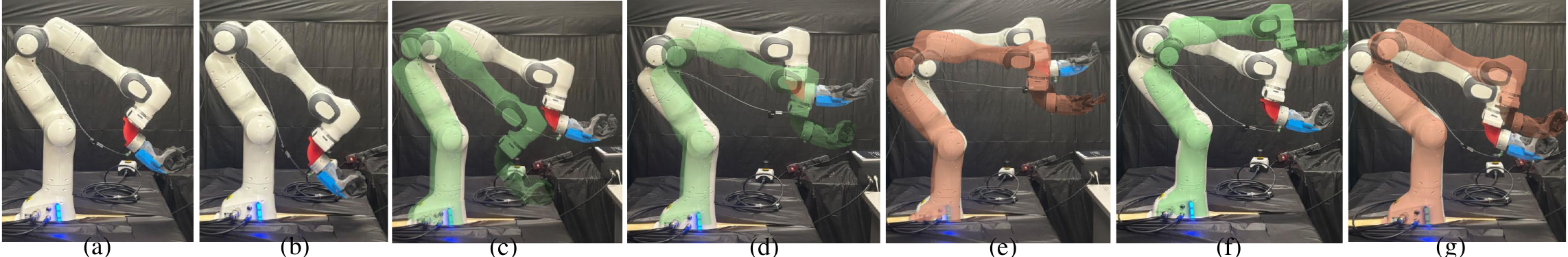}  
    \caption{Series snapshots of robot manipulation trajectory for completing tossing ball tasks. The robot moved into a downward position to accelerate from the position to the next throw/ ball release point. In green shadowed positions, we showed that the robot accelerates into the generated position. On the contrary, in the red shadowed position, we showed the de-acceleration of the robot's motion for smoother trajectory execution.}%
    \label{fig: snapshot}                                   
\end{figure*}

We use comprehensive experiments to evaluate our proposed algorithm for ball-tossing and catching tasks. Specifically, our experimental results include the following: 1) our algorithm successfully realized the alignment of human-robots and completed the tossing ball manipulation tasks with superior quantitative performance; 2) the functionality and visualization of CycleVAE; 3) the advantages of human behavior transformer.  

\subsection{Successful and smooth aligned manipulation trajectory of human robots}
We first presented a series of snapshots of our robot manipulation trajectory for achieving the standard ball tossing and catching task (see figure~\ref{fig: snapshot}).  The robot started its motion by moving to a downward configuration, intelligently positioning itself to create the momentum for the next toss ball phase. Robot configurations represented with a green overlay show acceleration phases, where the robot actively increased its velocity to reach the planned release point effectively. We observed that the robotic hand opens up at the perfect release timestep to throw the ball in upwards direction almost similar to a human arm motion for tossing the ball. On the contrary, poses highlighted in red indicated deceleration phases, during which the robot slowly reduced its velocity to ensure a smooth and controlled transition between motion segments. The robotic manipulator finished its final motion at the ball intercepting point with deceleration for a smooth catch. This modulation of acceleration and deceleration contributed to the overall stability and precision of the tossing action, mitigated abrupt motion changes, and enhanced trajectory execution. More detailed illustration figures can be found in our online videos.

\begin{figure}                                     
    \centering                                             
    \includegraphics[width=\linewidth]{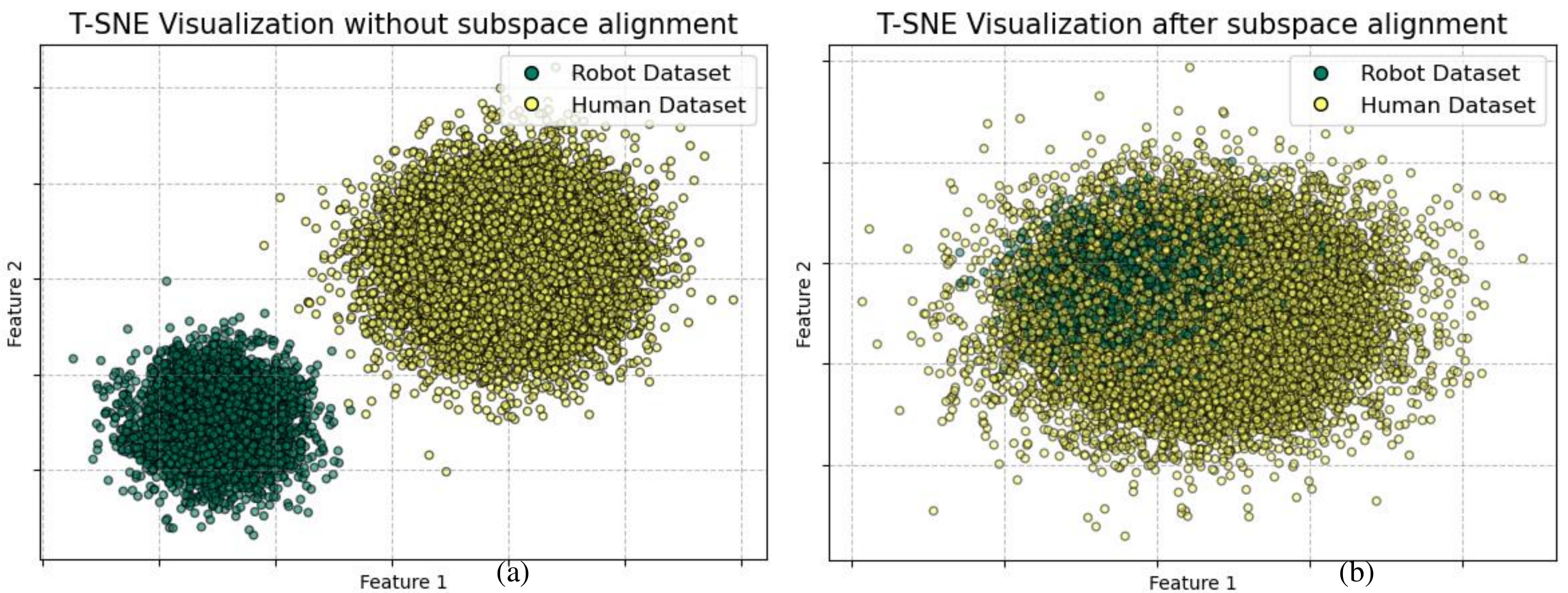}  
    \caption{T-SNE visualization to observe the effects of performing subspace alignment using of our proposed cycleVAE.}%
    \label{fig:tsne}                                   
\end{figure}

\subsection{Trade-offs comparisons to the SOTA algorithms}
\begin{figure*}                                     
    \centering                                             
    \includegraphics[width=\linewidth]{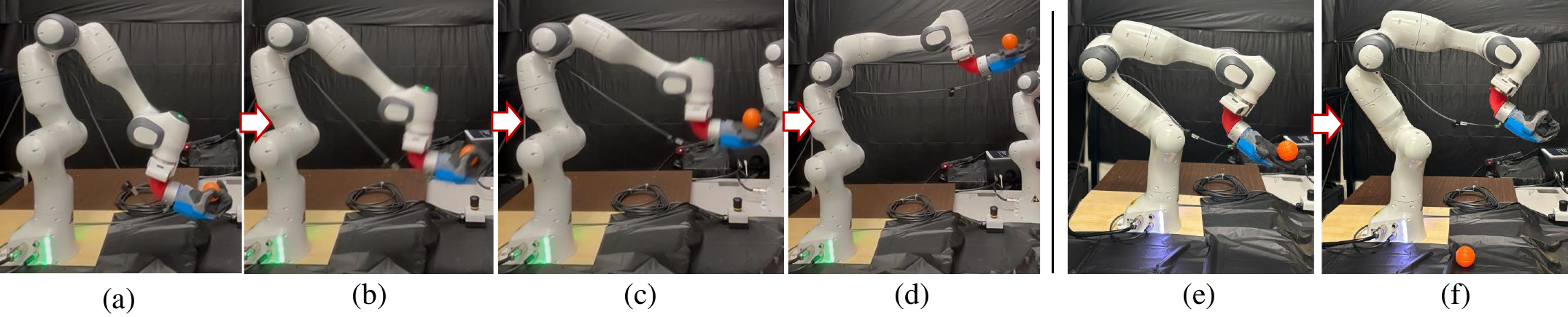}  
    \caption{(a)-(d) Failure to generate enough torques to release the ball at the exact point of release. (e)-(f) Failure to catch the thrown ball with accurate grasping configuration and ball falls on the table top.}%
    \label{fig: failure}                                   
\end{figure*}

We started by first illustrating several failure cases of our prior SOTA algorithms~\cite{Dastider_Damon_IROS_2023}. In figure~\ref{fig: failure}(a)-(d), we observed that DAMON could not produce enough torque values required for tossing the ball in the upward motion. Further, we noticed that the hand-closing motion was not perfectly synchronized, which led to the ball falling on the tabletop (see figure~\ref{fig: failure}(e)-(f)). 
With the above promising illustration comparison, we next comprehensively tested our proposed algorithm on $50$ different trials with two different size and weights of circular-shaped objects. Meanwhile, for a fair comparison, we also tested the other SOTA algorithms. Here, we mainly compared three quantitative performance metrics: 1) success ratio, 2) smoothness of the generated robotic trajectory, and 3) motion trajectory generation time. We summarized our results in Table~\ref{Table metrics comparsion}. From our averaged results, we observed that our proposed methods achieved all the best of three metrics under the simple light and smaller ball task. For the more practical tennis ball, our proposed method still achieved the highest success ratio and fastest trajectory generation speed. It is noted that the smoothness of our generated trajectory for the tennis ball was a little worse compared to the Vid2Robot~\cite{jain2024vid2robot}. It can be explained that Vid2Robot functions with finer pixel-level image information, while our method worked with only sparse $3D$ object information without including the object mesh information or geometrical shapes. In general, our method showed great advantages for completing tossing-ball tasks, which has implications for exploring domain alignment to achieve robotic tasks.

\begin{table}[]
\begin{threeparttable}
\begin{tabular}{|ccc|ccc|}
\hline
\multicolumn{1}{|c|}{Metric}           & \multicolumn{2}{|c|}{\begin{tabular}[c]{@{}c@{}}Success Ratio, \\ $R_{success}$(\%)\end{tabular}} & \multicolumn{2}{c|}{\begin{tabular}[c]{@{}c@{}}Smoothness, $S_r$\end{tabular}} & \begin{tabular}[c]{@{}c@{}}Motion Gen. \\ Time, $T_{gen}$\end{tabular} \\ \hline
\multicolumn{1}{|c|}{Method}           & \multicolumn{1}{c|}{A}                                      & B                                      & \multicolumn{1}{c|}{A}                       & \multicolumn{1}{c|}{B}                       & Avg.(sec)                                                               \\ \hline
\multicolumn{1}{|c|}{\textbf{Ours}}             & \multicolumn{1}{c|}{\textbf{91.2}}                                   & \textbf{86.3}                                   & \multicolumn{1}{c|}{\textbf{1.27}}                    & \multicolumn{1}{c|}{2.21}                    & \textbf{1.60}                                                                    \\ \hline
\multicolumn{1}{|c|}{DAMON}            & \multicolumn{1}{c|}{35.1}                                   & 29.7                                   & \multicolumn{1}{c|}{4.65}                    & \multicolumn{1}{c|}{4.95}                    & 6.31                                                                    \\ \hline
\multicolumn{1}{|c|}{Dynamic-RRT*}     & \multicolumn{1}{c|}{27.1}                                   & 15.2                                   & \multicolumn{1}{c|}{8.58}                    & \multicolumn{1}{c|}{9.67}                    & 9.12                                                                    \\ \hline
\multicolumn{1}{|c|}{Decision Trans.}  & \multicolumn{1}{c|}{79.3}                                   & 68.5                                   & \multicolumn{1}{c|}{5.41}                    & \multicolumn{1}{c|}{4.79}                    & 4.95                                                                    \\ \hline
\multicolumn{1}{|c|}{Vid2Robot}        & \multicolumn{1}{c|}{82.5}                                   & 76.7                                   & \multicolumn{1}{c|}{2.25}                    & \multicolumn{1}{c|}{1.85}                    & 2.74                                                                    \\ \hline
\multicolumn{1}{|c|}{Imitation Learn.} & \multicolumn{1}{c|}{65.3}                                   & 61.5                                   & \multicolumn{1}{c|}{7.36}                    & \multicolumn{1}{c|}{8.79}                    & 5.13                                                                    \\ \hline
\end{tabular}
\begin{tablenotes}
\footnotesize
\item A: Light Ball with smaller Size  \quad B: Tennis Ball with bigger Size  
\end{tablenotes}
\end{threeparttable}
\captionsetup{font={small,rm}}
\caption{Evaluation metrics comparison between our proposed method with prior SOTA methods on two experimental tasks. }
\label{Table metrics comparsion}
\end{table}

\subsection{Cycle-VAE aligns the lower dimensional human behavior manifold to the robotic manipulation manifold}
Having established the superior performance of our proposed algorithm for human-robotic domain alignment to achieve the tossing ball tasks, we zoomed two key components, CycleVAE and human behavior transformer (see next section). According to our method (see method section B), the proposed CycleVAE aims to align two disjoined human and robotic domains at latent space. To illustrate this functionality, we next applied the well-known T-SNE~\cite{tsne} for the visualization of both high-dimensional space (before alignment) and lower-dimensional latent space (after alignment). In figure~\ref{fig:tsne}, we observed that before the alignment, the human demonstration manifold and robotic trajectory manifold were almost disjoint from each other, while after the alignment, those two manifolds were mixed. To quantitatively measure the mixture level of data from two dissimilar domains, we also utilized the values of MMD distance between two distributions. Before alignment, we measured $L_{MMD} = 7.78$, which was also apparent in the figure~\ref{fig:tsne}; after performing bidirectional alignment, we calculated $L_{MMD} = 1.06$, which consistently showed the reduction gap via aligning the domain representation in learned subspaces.    

\subsection{Human behavior transformers expedite and simplify the behavior synthesis}
The second essential component in our algorithm is the human behavior transformer. Broadly speaking, we trained a causal transformer to predict the next step demonstration behavior, which can capture the internal dynamics of the human expert's behavior. Therefore, we next did another set of experiments by comparing the results using human expert demonstration with our well-trained human behavior transformer. We observed that our human transformer almost achieved the same success ratio compared to the human expert ($R_\text{success}$: expert: 91.2\% v.s. transformer: 89.4\%), indicating our casual human behavior successfully models human expert's behavior dynamics. Second, we compared the smoothness. It is believed that human experts always perform the feasible manipulation trajectory in a smooth and continuous way ($S_r$ = 1.27). However, our human behavior transformer also worked in a relative smooth way ($S_r$ = 1.53), especially when it also outperformed SOTA robotic trajectory planning algorithms in Table I. Last, the human transformer generated the trajectory faster than the human expert. This observation has future implications for using a transformer to replace human experts. Further, with the accelerated GPU and hardware platforms, the inference speed of the causal transformer will be even expedited, which may pave the way for large human behavior transformers to model various human behaviors simultaneously. To this end, Table \ref{Table II} summarized our comparison results.
\begin{table}[H]
\centering
\begin{tabular}{|c|ccc|}
\hline
Method         & \multicolumn{1}{c|}{$R_{success}$} & \multicolumn{1}{c|}{$S_r$} & $T_{gen}$ \\ \hline
Expert Demons.   & \multicolumn{1}{c|}{91.2}          & \multicolumn{1}{c|}{1.27}  & 1.58      \\ \hline
Behavior. Trans. & \multicolumn{1}{c|}{89.4}          & \multicolumn{1}{c|}{1.53}  & 1.31      \\ \hline

\end{tabular}
\captionsetup{font={small,rm}}
\caption{Comparison between human expert demonstration and human behavior transformer}
\label{Table II}
\end{table}

\section{DISCUSSIONS AND CONCLUSIONS}
In our work, we propose a new bidirectional human-robot alignment method for fast synthesis of robotic trajectory under the guidance of a human expert. We illustrated our method using tossing ball manipulation tasks, where our method outperformed SOTA algorithms. Although the tossing ball manipulation task is relatively simple compared to other fancy real-life behaviors, e.g., dancing, it can be treated as many fundamental manipulation behaviors of practical robots. Future work will test our method on other manipulation tasks, such as stacking~\cite{dastider2024apex}. We also proposed a causal human behavior transformer to learn internal human behavior dynamics, which can serve as the generative model for the synthesis of human expert demonstrations. Meanwhile, our further analysis of human behavior transformer showed the future potential of guiding robotic trajectory completely without human demonstrations. Last, inspired by the multi-modality ability of the transformer, we envision developing pure transformer-based bidirectional alignment algorithms, i.e., uniform multimodal transformer, for aligning human motion and robotic motion using contrastive learning~\cite{contrastive}.

\bibliographystyle{IEEEtran}
\bibliography{Bibliography}

\end{document}